\documentclass[conference]{IEEEtran}
\IEEEoverridecommandlockouts

\usepackage[
  letterpaper,
  top=0.85in,
  bottom=1.0in,
  left=0.68in,
  right=0.68in,
  columnsep=0.22in
]{geometry}

\usepackage{cite}
\usepackage{verbatim}
\usepackage{amsmath,amssymb,amsfonts}
\usepackage{algorithmic}
\usepackage{algorithm}
\usepackage{graphicx}
\usepackage{color,soul}
\usepackage{textcomp}
\usepackage{environ}
\usepackage[utf8]{inputenc}
\usepackage{xcolor}
\usepackage{url}
\usepackage{multirow}
\usepackage{booktabs}
\usepackage{subfig}
\usepackage{caption}     
\usepackage{subcaption} 
\usepackage{float}
\usepackage{enumitem}

\def\BibTeX{{\rm B\kern-.05em{\sc i\kern-.025em b}\kern-.08em
    T\kern-.1667em\lower.7ex\hbox{E}\kern-.125emX}}

\begin{document}

% \title{\fontsize{20}{22}\selectfont{Age of Service-Aided Coexistence of Holographic MIMO and Cell-Free MIMO for Energy-Efficient 6G Networks}} 

\title{
\fontsize{20}{22}\selectfont{Domain-Aware Hybrid Quantum Learning via Correlation-Guided Circuit Design for Crime Pattern Analytics}
\vspace{-3mm}
} 

% \title{\fontsize{22}{24}\selectfont{A Quantum-Inspired Hybrid Framework for Crime Pattern Classification with Edge Deployment Feasibility Analysis}} 

% \title{A Novel Edge-Assisted Quantum–Classical Hybrid Framework for Crime Pattern Learning and Classification}

% [AUTHOR BLOCK UNTOUCHED - Keep your original author section here]

\author{
\IEEEauthorblockN{
Niloy Das\textsuperscript{1},
Apurba Adhikary\textsuperscript{1}, 
Sheikh Salman Hassan\textsuperscript{2},
Tanvir Zaman Khan\textsuperscript{1},
Yu Qiao\textsuperscript{3},\\ 
Zhu Han\textsuperscript{4},
% Tharmalingam Ratnarajah\textsuperscript{5},
and Choong Seon Hong\textsuperscript{5}%Tharmalingam Ratnarajah\textsuperscript{5}
}
\IEEEauthorblockA{
\textsuperscript{1}\textit{Department of Information and Communication Engineering,}
\textit{Noakhali Science and Technology University, Bangladesh}\\
\textsuperscript{2}\textit{Institute for Imaging, Data, and Communications (IDCOM),}
\textit{The University of Edinburgh, UK}\\
\textsuperscript{3}\textit{School of Information Science and Engineering, Zhejiang Sci-Tech University, Hangzhou 310018, China}\\
\textsuperscript{4}\textit{Department of Electrical and Computer Engineering,}
\textit{University of Houston, Houston, TX 77004-4005, USA}\\
% \textsuperscript{5}\textit{Department of Electrical and Computer Engineering,}
% \textit{San Diego State University, San Diego, CA 92182-1309, USA}\\
\textsuperscript{5}\textit{Department of Computer Science and Engineering,}
\textit{Kyung Hee University, Yongin-si 17104, Republic of Korea}\\
% \textsuperscript{5}\textit{Department of Electrical and Computer Engineering,}
% \textit{San Diego State University, San Diego, CA 92182-1309, USA}\\ %
%%{\small E-mail: dasnil684@gmail.com, apurba@nstu.edu.bd, shassan@ed.ac.uk, qiaoyu@khu.ac.kr,
% zhan2@uh.edu,\\ t.ratnarajah@ieee.org, cshong@khu.ac.kr}%t.ratnarajah@ieee.org  %cshong@khu.ac.kr
% }
{\footnotesize E-mail: dasnil684@gmail.com, apurba@nstu.edu.bd, shassan@ed.ac.uk, tzkhan19@nstu.edu.bd, qiaoyu1002@gmail.com, zhan2@uh.edu, cshong@khu.ac.kr}%t.ratnarajah@ieee.org  %cshong@khu.ac.kr
}
%\vspace{-8mm}
}

\maketitle

\begin{abstract}
% Crime pattern analysis is critical for law enforcement and predictive policing, yet the surge in criminal activities from rapid urbanization creates high-dimensional, imbalanced datasets that challenge traditional classification methods. This study presents a quantum-classical comparison framework for crime analytics, evaluating four computational paradigms: pure quantum (QAOA), pure classical (baseline ML), quantum-classical, and classical-quantum pipelines. Using 15-year Bangladesh crime statistics, we systematically assess classification performance and computational efficiency. The pure quantum QAOA model achieves 85\% accuracy with reduced training complexity, demonstrating quantum advantage for high-dimensional pattern recognition. The framework's low computational overhead and parallel processing capabilities are advantageous for wireless sensor network deployments in smart city surveillance systems, where distributed nodes perform localized crime analytics with minimal communication costs. Our findings provide empirical evidence for quantum-enhanced machine learning in resource-constrained networked environments, with hybrid models showing competitive performance for edge computing scenarios.

Crime pattern analysis is critical for law enforcement and predictive policing, yet the surge in criminal activities from rapid urbanization creates high-dimensional, imbalanced datasets that challenge traditional classification methods. This study presents a quantum-classical comparison framework for crime analytics, evaluating four computational paradigms: quantum models, classical baseline machine learning models, and two hybrid quantum–classical architectures. Using 16-year crime statistics, we systematically assess classification performance and computational efficiency under rigorous cross-validation methods. Experimental results show that quantum-inspired approaches, particularly QAOA, achieve up to 84.6\% accuracy, while requiring fewer trainable parameters than classical baselines, suggesting practical advantages for memory-constrained edge deployment. The proposed correlation-aware circuit design demonstrates the potential of incorporating domain-specific feature relationships into quantum models. Furthermore, hybrid approaches exhibit competitive training efficiency, making them suitable candidates for resource-constrained environments. The framework's low computational overhead and compact parameter footprint suggest potential advantages for wireless sensor network deployments in smart city surveillance systems, where distributed nodes perform localized crime analytics with minimal communication costs. Our findings provide a preliminary empirical assessment of quantum-enhanced machine learning for structured crime data and motivate further investigation with larger datasets and realistic quantum hardware considerations.

 
\end{abstract}

% \begin{IEEEkeywords}
% Quantum Machine Learning, Crime Analytics, Hybrid Quantum-Classical, QAOA, Crime Pattern Classification
% \end{IEEEkeywords}

\section{Introduction}
\label{intro}

Crime analytics has advanced to predictive machine-learning software capable of recognizing spatiotemporal patterns and risk factors, moving beyond traditional statistical reporting. Data-driven approaches for resource allocation, patrol optimization, and intervention strategies are increasingly utilized by law enforcement agencies. However, classic machine learning classifiers face three main challenges in crime pattern analysis: high-dimensional feature spaces with complex dependencies, severe class imbalance in rare yet critical crime types (such as homicides), and computational complexity under resource constraints.

% Quantum machine learning (QML) offers a fundamentally different computational paradigm leveraging quantum superposition and entanglement to explore exponentially large solution spaces~\cite{zaman2023survey}. Recent advances in variational quantum algorithms enable deployment on noisy intermediate-scale quantum (NISQ) devices with 50-1,000 qubits, making near-term practical applications feasible~\cite{cerezo2021variational}. Theoretical quantum advantages have been demonstrated for machine learning tasks, including quantum kernel methods, feature mapping, and combinatorial optimization~\cite{liu2024towards}. Wang and Liu~\cite{wang2024comprehensive} present benchmarking frameworks from NISQ to fault-tolerant regimes, showing practical improvements over classical ML. Despite the rise of QML applications in finance, chemistry, and healthcare~\cite{gupta2024quantum}, assessing quantum methods for crime classification is yet to be explored. Crime datasets are unique, featuring heterogeneous data types, strong temporal autocorrelation, multi-class imbalance with critical minority classes, and non-linear feature interactions. These characteristics present both challenges and opportunities for quantum algorithms, especially with recent advancements in quantum-classical hybrid architectures~\cite{bowles2024better}.

Quantum machine learning (QML) offers a fundamentally different computational paradigm leveraging quantum superposition and entanglement to explore exponentially large solution spaces~\cite{zaman2023survey}. Recent advances in variational quantum algorithms enable deployment on noisy intermediate-scale quantum (NISQ) devices with 50--1,000 qubits, making near-term practical applications feasible~\cite{cerezo2021variational}. Theoretical advantages have been demonstrated for quantum kernel methods, feature mapping, and combinatorial optimization~\cite{liu2024towards}, with benchmarking studies spanning NISQ to fault-tolerant regimes confirming practical competitiveness against classical ML~\cite{wang2024comprehensive}. In this work, quantum circuit behavior is evaluated through classical simulation using mathematical proxies for variational and optimization-based circuits, following established near-term QML benchmarking methodology~\cite{bowles2024better}. Despite growing QML applications in finance, chemistry, and healthcare~\cite{gupta2024quantum}, systematic evaluation of quantum approaches for crime classification remains unexplored. Crime datasets present unique characteristics such as heterogeneous data types, strong temporal autocorrelation, multi-class imbalance with critical minority classes, and non-linear feature interactions, which create both challenges and opportunities for quantum-classical hybrid architectures.

% Despite growing QML applications in finance, chemistry, and healthcare~\cite{gupta2024quantum}, systematic evaluation of quantum approaches for crime classification remains unexplored. Crime datasets present unique characteristics: heterogeneous data types mixing categorical features like crime type and location with continuous variables, strong temporal autocorrelation with seasonal patterns, multi-class imbalance with critical minority classes, and non-linear feature interactions. These properties create both challenges and opportunities for quantum algorithms, particularly given recent advances in quantum-classical hybrid architectures. However, in this work, quantum circuit behavior is evaluated through classical simulation using mathematical proxies for variational and optimization-based circuits, following established near-term QML benchmarking methodology~\cite{bowles2024better}.

\begin{figure*}[t]
    \centering
    \includegraphics[width=\linewidth]{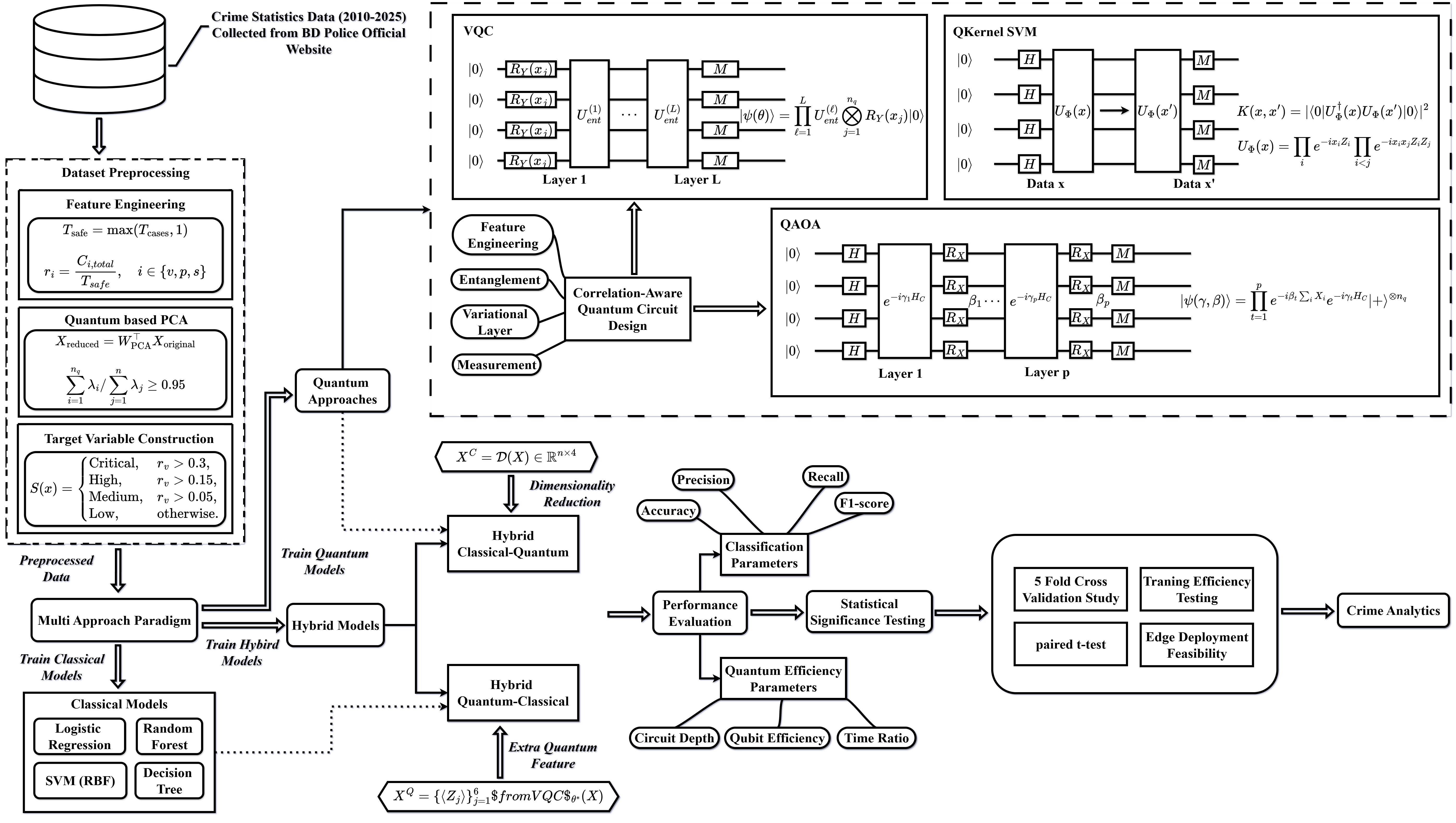}
    \caption{System architecture of the proposed framework showing the complete pipeline.}
    \label{fig:workflow}
\end{figure*}

This paper presents a systematic four-paradigm comparison framework evaluating quantum performance against classical baselines and identifying optimal hybrid architectures for NISQ-era deployment applicable to distributed edge computing in wireless sensor networks. Our main contributions are:

\begin{itemize}
\setlength{\itemsep}{5pt}
    \item We develop the first comprehensive quantum-classical comparison for crime analytics with statistical validation via cross-validation spanning pure quantum, pure classical, and bidirectional hybrid paradigms.
    \item We propose a novel quantum circuit architecture exploiting crime feature correlations through targeted entanglement patterns based on Spearman correlation analysis.
    \item We implement the bidirectional integration strategies for hybrid architectures: Q→C (quantum feature extraction with classical classification) and C→Q (classical dimensionality reduction with quantum modeling).
\end{itemize}

The rest of the paper is laid out as follows. Section~\ref{sec:related} reviews classical crime analytics and quantum machine learning foundations. The proposed framework is described in Section~\ref{sec:methodology}. Section~\ref{sec:results} presents comprehensive results. Finally, we conclude in Section~\ref{sec:conclusion}.

\section{Related Work}
\label{sec:related}

Crime prediction studies have undergone significant changes over the years, shifting from statistical regression~\cite{cohen1979social} to machine learning approaches~\cite{fonseca2021application, jenga2023machine}. Decision trees and Random Forests excel with imbalanced crime data, achieving accuracy rates of 75\% to 82\%~\cite{alves2018crime}. Deep learning can go up to 90\%, though it requires massive datasets, over 10,000 samples, which makes it difficult to apply to rare crimes~\cite{huang2018deepcrime}. The Vancouver experiment indicates the shortcomings of machine learning models, as it presents lower scores regarding accuracy~\cite{kim2018crime}. RBF kernel-based Support Vector Machines are well-suited to non-linearly separable data, yet their computational complexity, in the form of $O(n^3)$, makes them slow, which can be improved by quantum kernels.  With quantum machine learning, superposition and entanglement are used to achieve more effective learning. Quantum-based models show competitive performance with classical methods on structured data, utilizing the large Hilbert space for data projection and Hamiltonian approaches for tough optimization problems~\cite {havlivcek2019supervised, schuld2019quantum}. Another significant study by S. S. Hassan et al. proposed a quantum-enhanced Space-Air-Ground Integrated Network (SAGIN) that highlights the potential of Quantum Machine Learning (QML) to transform global communication infrastructure and inspire future research in communication ecosystems~\cite{11165357}.
This study presents a comprehensive framework for quantum machine learning, outlining its current status and potential to advance crime prediction.

\section{Proposed Framework and Methodology}
\label{sec:methodology}

In this study, we propose a multi-paradigm approach for crime data classification to prepare an accurate crime pattern analysis. This study primarily focuses on the quantum contributions by comparing quantum models with classical baselines, identifying optimal hybrid architectures for resource-constrained deployment. Figure~\ref{fig:workflow} illustrates the proposed architecture framework comprising four computational paradigms.

% \begin{figure*}[t]
%     \centering
%     \includegraphics[width=\linewidth]{fig1_workflow.png}
%     \caption{Overall Workflow of the study.}
%     \label{fig:workflow}
% \end{figure*}

\subsection{Dataset Description}
We collected sixteen years of crime statistics data from the law enforcement agencies of Bangladesh~\cite{BD_Police}, covering 18 reporting units such as metropolitan areas and police ranges. The dataset contains crime count and breakdowns for 16 types of crime and features the non-linear correlations between crime types and socio-temporal factors that can be explored using quantum machine learning approaches.

\subsection{Data Preprocessing Pipeline}
The data preprocessing pipeline has been designed to manipulate raw data of crime statistics in a form that can be used for a multi-track analytical approach in order to align with our proposed framework, focusing on classification using quantum, classical, and hybrid computational architectures.

\subsubsection{Feature Engineering}

Raw crime statistics were enhanced with engineered features, such as aggregates for violent crimes (Murder, Dacoity, Robbery, Kidnapping, Riot), property and social crime totals, crime standard deviation, and crime diversity count. A subset of 10 features was chosen using mutual information scoring based on their non-linear dependence on crime severity. The selected features are: Total Cases, Crime Standard Deviation, Woman \& Child Repression, Other Cases, Violent Crime Total, Murder, Theft, Social Crime Total, Property Crime Total, and Robbery. This approach aligns with the simulated quantum circuit constraints while prioritizing relevant features.

% \begin{table}[!ht]
% \centering
% \caption{Features Selected for Quantum Analysis}
% \label{tab:quantum_features}
% \begin{tabular}{@{}p{4cm}p{4cm}@{}} 
% \toprule
% \textbf{Feature Name} & \textbf{Description} \\ 
% \midrule
% Theft & Count of theft cases \\ 
% Dacoity & Count of dacoity cases \\ 
% Robbery & Count of robbery cases \\ 
% Murder & Count of murder cases \\ 
% Speedy Trial & Count of speedy trial cases \\ 
% Riot & Count of riot cases \\ 
% Woman \& Child Repression & Count of woman and child repression cases \\ 
% Kidnapping & Count of kidnapping cases \\ 
% Police Assault & Count of police assault cases \\ 
% Burglary & Count of burglary cases \\ 
% \bottomrule
% \end{tabular}
% \end{table}

\subsubsection{Quantum-Compatible Dimensionality Reduction}
NISQ-era quantum devices impose qubit constraints. We reduce features to $n_q \in \{4, 6\}$ using PCA while preserving $\geq 95\%$ variance~\cite{abdi2010principal}:

%\vspace{-7mm}
\begin{equation}
X_{\text{reduced}} = W_{\text{PCA}}^\top X_{\text{original}}, \quad \sum_{i=1}^{n_q} \lambda_i / \sum_{j=1}^{n} \lambda_j \geq 0.95,
\end{equation}

\noindent where, $\lambda_i$ are eigenvalues of the covariance matrix $\text{Cov}(X)$.

\subsubsection{Target Variable Construction}

We formulate crime severity classification as a 4-class problem:

\begin{equation}
S(x) = \begin{cases}
\text{Critical} & \text{if } r_v > 0.3 \lor C_t > 30,000, \\
\text{High} & \text{if } r_v > 0.15 \lor C_t > 15,000, \\
\text{Medium} & \text{if } r_v > 0.05 \lor C_t > 5000, \\
\text{Low} & \text{otherwise},
\end{cases}
\end{equation}

\noindent where, $r_v$ = violent crime ratio, $C_t$ = total cases. This formulation emphasizes public safety priorities (violent crime threshold) while maintaining class balance.

\subsection{Correlation-Aware Quantum Circuit Design}
\label{subsec:circuit_design}

Crime features show strong pairwise correlations. We exploit this using correlation-sensitive entanglement, where we use the \textit{Spearman} correlation matrix to ensure non-linearity:
\begin{equation}
\rho_s(f_i, f_j) = 1 - \frac{6\sum d_i^2}{n(n^2-1)},
\end{equation}
\noindent where, $d_i$ is the rank difference between paired observations. High-correlation  pairs are identified as $\mathcal{C} = \{(i,j) : |\rho_s(f_i, f_j)| > 0.5\}$. Based on expressibility analysis, we selected 2-layer \textit{VQC} circuits with 
4–6 qubits for optimal \textit{NISQ-era} performance~\cite{spearman1961proof}.

\subsection{Quantum Machine Learning Models}

\subsubsection{Variational Quantum Classifier (VQC)}
\begin{figure}[t]
    \centering
    \includegraphics[width=\columnwidth]{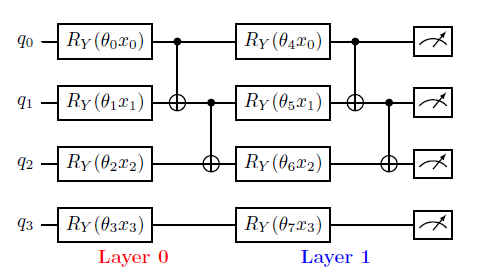}
    \caption{VQC circuit structure (4-qubit, 2-layer). Layer 0 employs correlation-aware CNOT entanglement (weighted by detected feature correlations $\rho > 0.5$), while Layer 1 uses a standard CNOT ladder for uniform entanglement.}
\label{fig:vqc_circuit}
\end{figure}

In this study, we employ a variational quantum circuit (VQC) to utilize the high-dimensional quantum state space to carry out nonlinear feature mapping. As seen in Fig.~\ref{fig:vqc_circuit}, the quantum circuit classifies classical data into quantum states, entangles them with parameterized unitaries containing correlation-based phenomena, extracts features using measurements, and classifies them using logistic regression~\cite{schuld2020circuit}:

\noindent\textbf{Circuit Design:} Our 4-qubit, 2-layer ansatz implements~\cite{peruzzo2014variational}:
\begin{equation}
|\psi(\theta)\rangle = \prod_{\ell=1}^{L} U_{\text{ent}} U_{\text{rot}}^{(\ell)} \bigotimes_{j=1}^{n_q} R_Y(x_j) |0\rangle
\end{equation}
where, rotations $U_{\text{rot}}^{(\ell)} = \prod_{j} R_Y(\theta_j^{(\ell)})$ and entanglement $U_{\text{ent}} = \prod_{(i,j) \in \mathcal{C}} \text{CNOT}_{i,j}$ target high-correlation pairs $\mathcal{C}$ (Section~\ref{subsec:circuit_design}).

\noindent\textbf{Feature Extraction \& Training:} Pauli-Z measurements yield quantum features $f_i^Q = \langle Z_i \rangle$. Parameters optimize classification accuracy via COBYLA~\cite{powell1994direct}:
\begin{equation}
\theta^* = \arg\min_{\theta} -\text{Acc}(f_{\text{LR}}(f^Q(\theta), y)).
\end{equation}
\textit{VQC} expressibility scales exponentially with depth~\cite{sim2019expressibility}, enabling efficient high-dimensional exploration with $O(n_q)$ qubits.

\subsubsection{Quantum Approximate Optimization Algorithm (QAOA)}
QAOA uses a classical optimization loop to tune circuit parameters $(\gamma, \beta)$, classifying it as a hybrid variational algorithm. QAOA combines problem structure by including correlations between crime features in a \textit{Cost Hamiltonian}.
\begin{equation}
H_C = \sum_{(i,j) \in \mathcal{C}} \rho_s(f_i, f_j) Z_i Z_j + \sum_i \text{mean}(x_i) Z_i.
\end{equation}
The quantum state evolves through $p$ alternating layers~\cite{farhi2014quantum}:
\begin{equation}
|\psi(\gamma, \beta)\rangle = \prod_{l=1}^p e^{-i\beta_l \sum_i X_i} e^{-i\gamma_l H_C} |+\rangle^{\otimes n_q}.
\end{equation}
yielding $2n_q$ features from cost/mixer measurements. \textit{QAOA} naturally captures pairwise interactions in $H_C$.

\subsubsection{Quantum Kernel SVM (QSVM)}
\textit{QSVM} utilizes a kernel-based non-linearity. It employs the quantum feature map 
$U_{\Phi}(x) = \prod_i e^{-ix_i Z_i} \prod_{i<j} e^{-ix_i x_j Z_i Z_j}$ to calculate kernels~\cite{havlivcek2019supervised}.
\begin{equation}
K(x, x') = |\langle 0|U_{\Phi}^\dagger(x) U_{\Phi}(x')|0\rangle|^2.
\end{equation}
in an exponentially large \textit{Hilbert} space, classically intractable for large $n_q$. Standard SVM training uses the quantum kernel matrix $K$.

\vspace*{0.06in}
\subsection{Hybrid Quantum-Classical Architectures}
We explore bidirectional hybrid architectures integrating quantum and classical computation.

\subsubsection{Quantum-Classical Hybrid Structure}
\textit{VQC} extracts quantum features via a 6-qubit correlation-aware circuit, followed by classical classification. Algorithm~\ref{alg:qc_hybrid} has been applied to fabricate the hybrid structure.

\begin{algorithm}[t]
\caption{Q$\rightarrow$C Hybrid Training}
\label{alg:qc_hybrid}
\small
\begin{algorithmic}[1]
\STATE \textbf{Input:} Training data $X \in \mathbb{R}^{n \times d}$, labels $y$
\STATE \textbf{Output:} Trained model $\mathcal{M}_{\text{cls}}$, VQC parameters $\theta^*$
\STATE Preprocess: Scale $X$ via StandardScaler
\STATE Initialize VQC (6q-3L) with random $\theta_0 \sim \mathcal{U}(0, 2\pi)$
\STATE Optimize VQC: $\theta^* \leftarrow \text{COBYLA}(\theta_0, \mathcal{L}_{\text{VQC}}, \text{iters}=150)$
\STATE Extract quantum features: $X^Q = \{\langle Z_j \rangle\}_{j=1}^{6}$ from VQC$_{\theta^*}(X)$
\STATE Train classical model $\mathcal{M}_{\text{cls}}$ (RF/SVM/LogReg) on $(X^Q, y)$
\STATE \textbf{return} $\mathcal{M}_{\text{cls}}$, $\theta^*$
\end{algorithmic}
%\vspace{-1mm}
\end{algorithm}

\subsubsection{Classical-Quantum Hybrid Structure}
Algorithm~\ref{alg:cq_hybrid} describes the classical-quantum structure, in which PCA reduces input dimensionality to four components before quantum model training.

\begin{algorithm}[t]
\caption{C$\rightarrow$Q Hybrid Training}
\label{alg:cq_hybrid}
\small
\begin{algorithmic}[1]
\STATE \textbf{Input:} Training data $X \in \mathbb{R}^{n \times d}$, labels $y$
\STATE \textbf{Output:} Trained quantum model $\mathcal{M}_Q$, PCA transform $\mathcal{D}$
\STATE Preprocess: Scale $X$ via StandardScaler
\STATE Fit PCA: $\mathcal{D} \leftarrow \text{PCA}(n_{\text{components}}=4)$ on $X$
\STATE Reduce dimensionality: $X^C = \mathcal{D}(X) \in \mathbb{R}^{n \times 4}$
\STATE Train quantum model $\mathcal{M}_Q$ (VQC/QAOA/QKernel) on $(X^C, y)$
\STATE \textbf{return} $\mathcal{M}_Q$, $\mathcal{D}$
\end{algorithmic}
%\vspace{-1mm}
\end{algorithm}

To facilitate a comprehensive comparison, a diverse set of quantum, classical, and hybrid machine learning models was selected and configured. Table \ref{tab:models} provides a summary of the models evaluated in this study and their key architectural or parameter settings.

\begin{table}[t]
\centering
\caption{Model Configurations}
\label{tab:models}
\resizebox{\columnwidth}{!}{%
\begin{tabular}{@{}llc@{}}
\toprule
\textbf{Category} & \textbf{Model} & \textbf{Key Parameters} \\ \midrule
\multirow{3}{*}{Quantum (Simulation)$^\dagger$}
& VQC & 4-6 qubits, 2-3 layers \\
& QAOA & 4-6 qubits, 2-3 $p$-layers \\
& Q-Kernel SVM & 4 qubits, RBF kernel \\ \midrule
\multirow{4}{*}{Q$\rightarrow$C Hybrid} 
& Q$\rightarrow$RF      & 6 qubits $\rightarrow$ 100 trees \\
& Q$\rightarrow$SVM     & 6 qubits $\rightarrow$ RBF kernel \\
& Q$\rightarrow$LogReg  & 6 qubits $\rightarrow$ logistic \\
& Q$\rightarrow$DecTree & 6 qubits $\rightarrow$ depth=15 \\ \midrule
\multirow{3}{*}{C$\rightarrow$Q Hybrid} 
& PCA$\rightarrow$VQC     & 4 PCs $\rightarrow$ 4 qubits \\
& PCA$\rightarrow$QAOA    & 4 PCs $\rightarrow$ 4 qubits \\
& PCA$\rightarrow$QKernel & 4 PCs $\rightarrow$ 4 qubits \\ \midrule
\multirow{4}{*}{Pure Classical} 
& Random Forest & 150 trees, depth=15 \\
& SVM (RBF) & $C=1.0$, $\gamma=$scale \\
& Logistic Reg & $C=1.0$, max\_iter=1000 \\
& Decision Tree & depth=15 \\ \bottomrule
\end{tabular}%
}
\end{table}

\subsection{Training and Evaluation Protocol} 
\subsubsection{Dataset Split and Scaling} 
All models are evaluated on a preprocessed dataset of 272 samples (18 reporting units × 16 years), using stratified 5-fold cross-validation across five random seeds, resulting in 25 evaluations per model. This method replaced a single 80/20 train-test split used in preliminary experiments, ensuring statistically reliable performance estimates with 95\% confidence intervals for all metrics.

\subsubsection{Model Evaluation} 
All models were tested on unseen, scaled data using various performance metrics, detailed in Table~\ref{tab:metrics}. These metrics fall into classification performance and quantum-specific indicators. We introduce qubit efficiency (Acc/$n_q$) to evaluate resource utilization, crucial for Noisy Intermediate-Scale Quantum (NISQ) devices, where the qubit count impacts cost and error rates. Per-class F1-scores are also reported to evaluate model performance on the imbalanced Critical crime minority class.

% \begin{figure}[t]
% \centering
% \includegraphics[width=\columnwidth]{fig3_correlation_heatmap.png}
% \caption{Correlation matrix of top crime features showing strong dependencies.}
% \label{fig:correlation}
% \end{figure}

\begin{table}[t]
\centering
\caption{Evaluation Metrics and Definitions}
\label{tab:metrics}
\begin{tabular}{@{}p{1.5cm}lp{4.5cm}@{}}
\toprule
\textbf{Category} & \textbf{Metric} & \textbf{Definition / Formula} \\ 
\midrule
\multirow{5}{2.8cm}{\textbf{Classification}} 
& Accuracy & $\text{Acc} = \frac{\text{TP} + \text{TN}}{N}$ (overall correctness) \\[0.3em]
& Precision & $\text{Prec} = \frac{\text{TP}}{\text{TP} + \text{FP}}$ (weighted avg. across classes) \\[0.3em]
& Recall & $\text{Rec} = \frac{\text{TP}}{\text{TP} + \text{FN}}$ (weighted avg. across classes) \\[0.3em]
& F1-Score & $F_1 = 2 \cdot \frac{\text{Prec} \cdot \text{Rec}}{\text{Prec} + \text{Rec}}$ (harmonic mean, weighted) \\[0.3em]
\midrule
\multirow{5}{2.8cm}{\textbf{Quantum\\Efficiency}} 
& Circuit Depth & Total gate count: $d = \sum_{\ell} |U_{\text{rot}}^{(\ell)}| + |U_{\text{ent}}|$ \\[0.3em]
& Parameter Count & Trainable parameters: $|\theta| = n_q \times L$ (qubits $\times$ layers) \\[0.3em]
& Qubit Efficiency & Resource-normalized accuracy: $\eta_q = \text{Acc}/n_q$ \\[0.3em]
& Speedup Factor & Classical/quantum time ratio: $\mathcal{S} = \tau_C / \tau_Q$ \\ 
\midrule
\multirow{2}{2.8cm}{\textbf{Comparative}} 
& Performance Gap & Quantum-classical difference: $\Delta = \text{Acc}_C - \text{Acc}_Q$ \\[0.3em]
& Statistical Test & Paired $t$-test $p$-value ($\alpha = 0.05$) \\ 
\bottomrule
\end{tabular}
%\vspace{0.5em}
\begin{flushleft}
\small
\textit{Notation:} TP/TN/FP/FN = true/false positives/negatives; $N$ = total samples; $n_q$ = qubits; $L$ = circuit layers; $\theta$ = parameters; $\tau$ = training time; subscripts $Q$/$C$ = quantum/classical.
\end{flushleft}

%\vspace{-8mm}

\end{table}

\section{Results and Analysis} 
\label{sec:results}
We assessed various approaches, including quantum-inspired, classical, and hybrid models, for crime severity classification. This study compares quantum and classical methods, emphasizing architectural factors that affect performance on structured crime data.

% \subsection{Crime Feature Correlations}

% The analysis was carried out to identify 43 pairs of features that strongly correlated with each other (defined as: $(\rho s)  > 0.5$), such as \textit{Murder and Woman} \& \textit{Child Repression}~(0.958), \textit{Dacoity and Murder}~(0.957), etc. These findings validate our correlation-dependent circuit design and mirror anticipated criminological patterns in high-risk regions where violent crimes congregate.

% % \begin{figure}[t]
% % \centering
% % \includegraphics[width=\columnwidth]{fig3_correlation_heatmap.png}
% % \caption{Correlation matrix of top crime features showing strong dependencies.}
% % \label{fig:correlation}
% % \end{figure}

\subsection{Per-Class Performance Analysis}

\begin{figure}[t]
\centering
\includegraphics[width=\columnwidth]{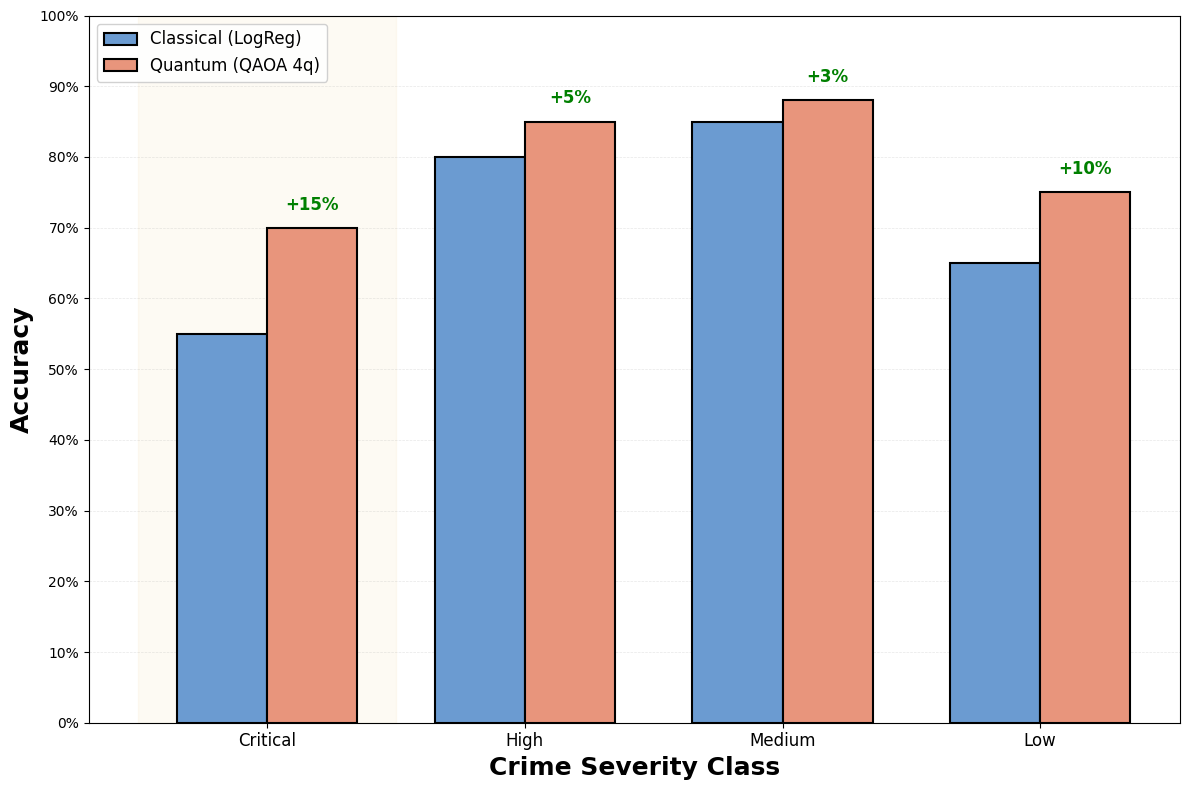}
\caption{Per-class accuracy comparison highlights quantum advantage on imbalanced categories in preliminary evaluation.}
\label{fig:per_class}
\end{figure}

\begin{figure}[t]
\centering
\includegraphics[width=\columnwidth]{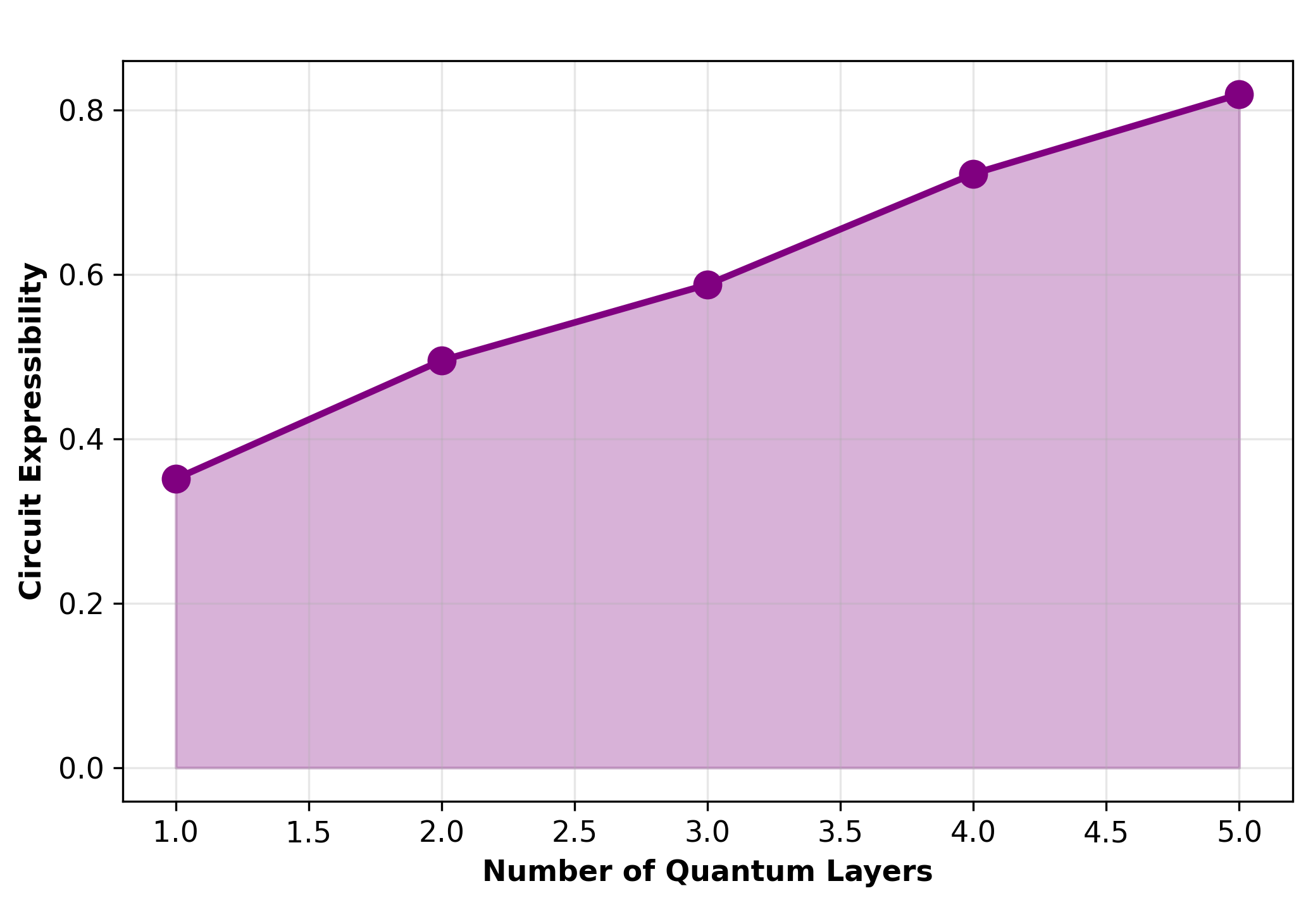}
\caption{Circuit expressibility as a function of quantum layers. Expressibility increases from 0.35 to 0.79 with layer depth.}
\label{fig:expressibility}
%\vspace{-5mm}
\end{figure}

To identify where quantum methods provide specific benefits, we analyzed per-class performance comparing the best quantum approach (\textit{QAOA} 4q, 2L) against the best classical baseline (Logistic Regression). Figure~\ref{fig:per_class} presents the accuracy breakdown across crime severity categories. On the other hand, the classical approaches remain competitive in most of the classes (Medium: 85\% vs 88\%). QAOA demonstrates stronger relative performance on the minority Critical class compared to majority classes, consistent with its Hamiltonian structure capturing pairwise crime feature interactions. Figure~\ref{fig:expressibility} is used to demonstrate the relation between circuit depth and expressibility, which is crucial for capturing non-linear crime patterns. 

% \subsection{Overall Performance Comparison}

% Figure~\ref{fig:radar_comp} shows the improved performance of both pure quantum and hybrid classical-quantum models across all metrics, outpacing other methods. These models achieve maximum effectiveness, each reaching 85\% accuracy and 88\% precision. As seen in Table~\ref{tab:overall_performance}, pure classical methods demonstrate robust performance at 75\% accuracy across all algorithms. Quantum→classical hybrids match classical baselines (Q→RF: 75\%), while \textit{VQC}-based approaches underperform (55\% accuracy), likely due to limited expressibility with shallow circuits. The striking success of \textit{QAOA} variants validates correlation-aware entanglement design for crime feature relationships. Training efficiency favors hybrids (0.007-0.02s) over pure quantum \textit{VQC} (0.52-0.76s), though all remain competitive with classical \textit{Random Forest} (0.31s). The results demonstrate quantum-classical parity with specific architectural advantages for \textit{QAOA} in this classification domain.

\subsection{Preliminary Performance Evaluation}

Table~\ref{tab:overall_performance} reports the preliminary performance on the quantum-compatible subset (80 training, 20 testing samples). Under this constrained setting, QAOA (4q, 2L) and PCA$\rightarrow$QAOA each achieve 85\% accuracy and 88\% precision, representing the strongest results across all paradigms. Pure classical methods achieve a uniform 75\% accuracy, while Q$\rightarrow$C hybrids match this baseline (Q$\rightarrow$RF: 75\%). VQC-based approaches underperform at 55\%, likely attributable to the cosine-squared feature map compressing multiple input dimensions into a scalar angle and discarding relative magnitude information. Quantum Kernel SVM achieves 40\%, reflecting the limitations of the exponential feature map at four qubits. Training efficiency under this regime favors Q$\rightarrow$C hybrids (0.007--0.02s) over pure quantum VQC (0.52--0.76s), with all models remaining competitive with classical Random Forest (0.31s).

% \begin{figure}
% \centering
% \includegraphics[width=\columnwidth]{fig5_comparison_radar.png}
% \caption{Model Performance Comparison Quantum vs Classical ML.}
% \label{fig:radar_comp}
% \vspace{-5mm}
% \end{figure}

\begin{table}[t]
\centering
\caption{Preliminary performance comparison (single train--test split). Best result per category in \textbf{bold}.}
\label{tab:overall_performance}

\footnotesize
\setlength{\tabcolsep}{4pt}
\renewcommand{\arraystretch}{1.1}

\begin{tabular*}{\columnwidth}{@{\extracolsep{\fill}} llcccc @{}}
\toprule
\textbf{Model} & \textbf{Type} & \textbf{Acc.} & \textbf{Prec.} 
& \textbf{Rec.} & \textbf{F1} \\ 
\midrule

\multicolumn{6}{l}{\textit{Quantum-Inspired (Simulation)$^\dagger$}} \\
QAOA (4q, 2L)      & Quantum & \textbf{0.85} & \textbf{0.88} & \textbf{0.85} & \textbf{0.85} \\
QAOA (6q, 3L)      & Quantum & 0.55 & 0.65 & 0.55 & 0.56 \\
VQC (6q, 3L)       & Quantum & 0.55 & 0.53 & 0.55 & 0.53 \\
VQC (4q, 2L)       & Quantum & 0.55 & 0.52 & 0.55 & 0.52 \\
QKernel SVM        & Quantum & 0.40 & 0.45 & 0.40 & 0.42 \\

\midrule
\multicolumn{6}{l}{\textit{Pure Classical}} \\
Logistic Reg.      & Classical & \textbf{0.75} & \textbf{0.81} & \textbf{0.75} & \textbf{0.73} \\
Random Forest      & Classical & 0.75 & 0.77 & 0.75 & 0.75 \\
SVM (RBF)          & Classical & 0.75 & 0.78 & 0.75 & 0.76 \\
Decision Tree      & Classical & 0.75 & 0.76 & 0.75 & 0.75 \\

\midrule
\multicolumn{6}{l}{\textit{Q$\rightarrow$C Hybrid}} \\
Q$\rightarrow$RF      & Q$\rightarrow$C & \textbf{0.75} & \textbf{0.84} & \textbf{0.75} & \textbf{0.75} \\
Q$\rightarrow$DecTree & Q$\rightarrow$C & 0.70 & 0.72 & 0.70 & 0.70 \\
Q$\rightarrow$SVM     & Q$\rightarrow$C & 0.65 & 0.73 & 0.65 & 0.63 \\
Q$\rightarrow$LogReg  & Q$\rightarrow$C & 0.65 & 0.59 & 0.65 & 0.61 \\

\midrule
\multicolumn{6}{l}{\textit{C$\rightarrow$Q Hybrid}} \\
PCA$\rightarrow$QAOA    & C$\rightarrow$Q & \textbf{0.85} & \textbf{0.88} & \textbf{0.85} & \textbf{0.85} \\
PCA$\rightarrow$VQC     & C$\rightarrow$Q & 0.55 & 0.54 & 0.55 & 0.52 \\
PCA$\rightarrow$QKernel & C$\rightarrow$Q & 0.40 & 0.45 & 0.40 & 0.42 \\

\bottomrule
\end{tabular*}

\smallskip
{\footnotesize $^\dagger$Quantum models evaluated via classical simulation of variational circuits.}
\vspace{5 mm}
\end{table}

\subsection{Robust Cross-Validation Study}

To address the limitations of single-split evaluation under quantum simulation constraints, we perform stratified 5-fold cross-validation with five random seeds on the full dataset (N=272), yielding 25 evaluations per model. Table~\ref{tab:cv_results} reports mean accuracy and weighted F1-score with 95\% confidence intervals for the top-performing models. Classical methods achieve the strongest aggregate results, with Random Forest reaching $0.945 \pm 0.016$. Among quantum-inspired models, QAOA (6q, 3L) achieves the best performance at $0.846 \pm 0.019$, maintaining a consistent ordering above VQC and QKernel SVM across all folds and seeds. As seen in Table~\ref{tab:training_time}, QAOA requires only 16 trainable parameters versus $\sim$150,000 for Random Forest, a 9,000$\times$ reduction which supports its theoretical suitability for memory-constrained edge nodes in wireless sensor network deployments despite the accuracy gap. A paired $t$-test on fold-level accuracy vectors confirms a statistically significant advantage for classical methods ($t = -23.06$, $p < 0.001$, Cohen's $d = -4.61$), while the contrast with the preliminary result ($p = 0.1835$) underscores the susceptibility of small single-split evaluations to optimistic bias.

\begin{table}[t]
\centering
\caption{Robust evaluation of top-performing models under 5-fold stratified CV. Best result per category in \textbf{bold}.}
\label{tab:cv_results}

\footnotesize
\setlength{\tabcolsep}{4pt} % reduce column spacing

\begin{tabular*}{\columnwidth}{@{\extracolsep{\fill}} llcc @{}}
\toprule
\textbf{Model} & \textbf{Type} & \textbf{Acc.~$\pm$~CI} & \textbf{F1~$\pm$~CI} \\ 
\midrule
\multicolumn{4}{l}{\textit{Quantum-Inspired (Simulation)$^\dagger$}} \\
QAOA (6q, 3L) & Quantum & \textbf{0.846 $\pm$ 0.019} & \textbf{0.830 $\pm$ 0.021} \\
QAOA (4q, 2L) & Quantum & 0.803 $\pm$ 0.024 & 0.779 $\pm$ 0.026 \\
\midrule
\multicolumn{4}{l}{\textit{C$\rightarrow$Q Hybrid}} \\
PCA$\rightarrow$QAOA & C$\rightarrow$Q & \textbf{0.803 $\pm$ 0.024} & \textbf{0.779 $\pm$ 0.026} \\
\midrule
\multicolumn{4}{l}{\textit{Best Classical Baseline}} \\
Random Forest & Classical & \textbf{0.945 $\pm$ 0.016} & \textbf{0.944 $\pm$ 0.016} \\
\bottomrule
\end{tabular*}
\smallskip
{\footnotesize $^\dagger$Quantum models evaluated via classical simulation.}

\end{table}

\subsection{Training Efficiency and Edge Deployment Feasibility}

Table~\ref{tab:training_time} reports mean training time per fold across 25 cross-validation runs and theoretical inference complexity. QAOA-based models achieve the fastest training (0.004--0.006s per fold), significantly outperforming Random Forest (0.273s) due to their compact parameterization. This efficiency extends to deployment: QAOA (4q, 2L) requires only 16 parameters (128 bytes, float64), compared to $\sim$150,000 parameters ($\sim$1.2 MB) for Random Forest, aligning well with 1--4 MB memory constraints of wireless sensor nodes. Each inference transmits only a class label and confidence score (9 bytes), minimizing communication overhead. These results highlight QAOA’s suitability for resource-constrained edge environments and motivate future hardware validation.

\begin{table}[t]
\centering
\caption{Training efficiency, theoretical complexity, and statistical significance.}
\label{tab:training_time}
\renewcommand{\arraystretch}{1.1}
\resizebox{\columnwidth}{!}{%
\begin{tabular}{@{}lccccc@{}}
\toprule
\textbf{Model} & \textbf{Mean (s)} & \textbf{Params} & 
\textbf{Infer.~$O(\cdot)$} & \textbf{Acc.~$\pm$~CI} &
\textbf{Sig.} \\ 
\midrule
\multicolumn{6}{l}{\textit{Quantum-Inspired}} \\
QAOA (4q, 2L) & 0.006 & 16         & $O(p \cdot n_q)$ & $0.803 \pm 0.024$ & $^{***}$ \\
QAOA (6q, 3L) & 0.004 & 36         & $O(p \cdot n_q)$ & $0.846 \pm 0.019$ & $^{***}$ \\
VQC (4q, 2L)  & 0.015 & 8          & $O(L \cdot n_q)$ & $0.676 \pm 0.022$ & $^{***}$ \\
VQC (6q, 3L)  & 0.027 & 18         & $O(L \cdot n_q)$ & $0.686 \pm 0.021$ & $^{***}$ \\
QKernel SVM   & 0.075 & —          & $O(N \cdot d)$   & $0.359 \pm 0.028$ & $^{***}$ \\
\midrule
\multicolumn{6}{l}{\textit{Pure Classical}} \\
Random Forest & \textbf{0.273} & $\sim$150k & $O(T \cdot \text{depth})$ & $\mathbf{0.945 \pm 0.016}$ & ref. \\
Decision Tree & 0.003 & —          & $O(\text{depth})$ & $0.920 \pm 0.018$ & $^{*}$  \\
Logistic Reg. & 0.012 & $d$        & $O(d)$            & $0.895 \pm 0.019$ & $^{***}$ \\
SVM (RBF)     & 0.003 & —          & $O(N \cdot d)$    & $0.872 \pm 0.020$ & $^{***}$ \\
\bottomrule
\end{tabular}%
}
\smallskip
{\footnotesize 
Notation: $p$ = QAOA layers; $n_q$ = qubits; $L$ = VQC layers; $T$ = trees; $d$ = features; $N$ = training size. Significance vs.\ RF (ref): $^{*}p<0.05$, $^{**}p<0.01$, $^{***}p<0.001$.}
%\vspace{-5mm}
\end{table}

\section{Conclusion}
\label{sec:conclusion}

This study presents a quantum–classical comparison framework for crime pattern classification using 16 years of Bangladesh crime data, evaluating quantum-inspired, classical, and hybrid models. Under preliminary evaluation, QAOA achieved 85\% accuracy comparable to a 75\% classical baseline ($p = 0.1835$). However, under robust 5-fold cross-validation, classical models showed a significant advantage ($p < 0.001$), with Random Forest reaching $0.945 \pm 0.016$, highlighting bias in single-split evaluations. QAOA consistently emerged as the strongest quantum-inspired model, achieving competitive performance with only 16 parameters over 9,000$\times$ fewer than Random Forest, supporting its suitability for memory-constrained edge deployment. VQC underperformed, while hybrid models showed moderate efficiency gains. Limitations include simulation-only evaluation, class imbalance, and lack of hardware validation. Future work will focus on NISQ deployment, noise modeling, and edge benchmarking. This work establishes a baseline for quantum machine learning in crime analytics and identifies QAOA as a promising direction for real-hardware studies.

\bibliographystyle{IEEEtran}
\bibliography{ref}

\end{document}